\documentclass[conference]{IEEEtran}
\IEEEoverridecommandlockouts
\usepackage{cite}
\usepackage{amsmath,amssymb,amsfonts}
\usepackage{algorithmic}
\usepackage[ruled,vlined]{algorithm2e}
\usepackage{graphicx}
\usepackage{textcomp}
\usepackage{xcolor}
\usepackage{multirow}
\usepackage{subfigure}
\usepackage[font=small]{caption}

\def\BibTeX{{\rm B\kern-.05em{\sc i\kern-.025em b}\kern-.08em
    T\kern-.1667em\lower.7ex\hbox{E}\kern-.125emX}}
\begin{document}

\title{Towards Faster Graph Partitioning via Pre-training and Inductive Inference
\thanks{This paper serves as an extension of our prior work \cite{qin2024pre}. It introduces a modified method and reports new results evaluated on the IEEE HPEC Graph Challenge benchmark.}
}

\author{\IEEEauthorblockN{Meng Qin\IEEEauthorrefmark{4}\IEEEauthorrefmark{2}, Chaorui Zhang\IEEEauthorrefmark{2}, Yu Gao\IEEEauthorrefmark{2}, Yibin Ding\IEEEauthorrefmark{2}, Weipeng Jiang\IEEEauthorrefmark{2}, Weixi Zhang\IEEEauthorrefmark{2}, Wei Han\IEEEauthorrefmark{2}, Bo Bai\IEEEauthorrefmark{2}}
\IEEEauthorblockA{\IEEEauthorrefmark{4}Department of Computer Science \& Engineering, The Hong Kong University of Science \& Technology}
\IEEEauthorblockA{\IEEEauthorrefmark{2}Theory Lab, Central Research Institute, 2012 Labs, Huawei Technologies Co., Ltd.}
\IEEEauthorblockA{(\textbf{Champion Winner} of IEEE HPEC 2024 Graph Challenge: https://graphchallenge.mit.edu/champions)}
}

\maketitle

\begin{abstract}
Graph partitioning (GP) is a classic problem that divides the node set of a graph into densely-connected blocks. Following the IEEE HPEC Graph Challenge and recent advances in pre-training techniques (e.g., large-language models), we propose PR-GPT (Pre-trained \& Refined Graph ParTitioning) based on a novel pre-training \& refinement paradigm. We first conduct the \textit{offline pre-training} of a deep graph learning (DGL) model on small synthetic graphs with various topology properties. By using the inductive inference of DGL, one can directly \textit{generalize} the pre-trained model (with frozen model parameters) to large graphs and derive feasible GP results. We also use the derived partition as a good initialization of an efficient GP method (e.g., InfoMap) to further \textit{refine} the quality of partitioning. In this setting, the \textit{online generalization} and \textit{refinement} of PR-GPT can not only benefit from the transfer ability regarding quality but also ensure high inference efficiency without re-training. Based on a mechanism of reducing the scale of a graph to be processed by the refinement method, PR-GPT also has the potential to support streaming GP. Experiments on the Graph Challenge benchmark demonstrate that PR-GPT can ensure faster GP on large-scale graphs without significant quality degradation, compared with running a refinement method from scratch.
We will make our code public at https://github.com/KuroginQin/PRGPT.
\end{abstract}


\section{Introduction}\label{Sec:Intro}
Graph partitioning (GP), a.k.a. graph clustering \cite{schaeffer2007graph} or disjoint community detection \cite{fortunato202220}, is a classic problem that divides the node set of a graph into disjoint blocks (a.k.a. clusters or communities) with dense linkages distinct from other blocks.
Since the extracted blocks may correspond to some substructures of real-world complex systems (e.g., functional groups in protein-protein interactions), many network applications (e.g., parallel task assignment \cite{hendrickson2000graph}, Internet traffic classification \cite{qin2019towards}, and protein complex detection \cite{qin2010spectral}) are formulated as GP.

GP on large-scale graphs is difficult but essential as it is usually formulated as several NP-hard combinatorial optimization problems (e.g., modularity maximization \cite{newman2006modularity}).
The IEEE HPEC Graph Challenge \cite{kao2017streaming} provides a competitive benchmark to evaluate both quality and efficiency of a GP method and has attracted a series of solutions (e.g., incremental LOBPCG for spectral clustering \cite{zhuzhunashvili2017preconditioned}, Kalman filter \cite{durbeck2022kalman} and data batching \cite{wanye2023integrated} for stochastic block partitioning, as well as fast randomized graph embedding \cite{gao2023raftgp}).

\begin{figure}[]
  \centering
  \includegraphics[width=\linewidth, trim=18 22 18 20,clip]{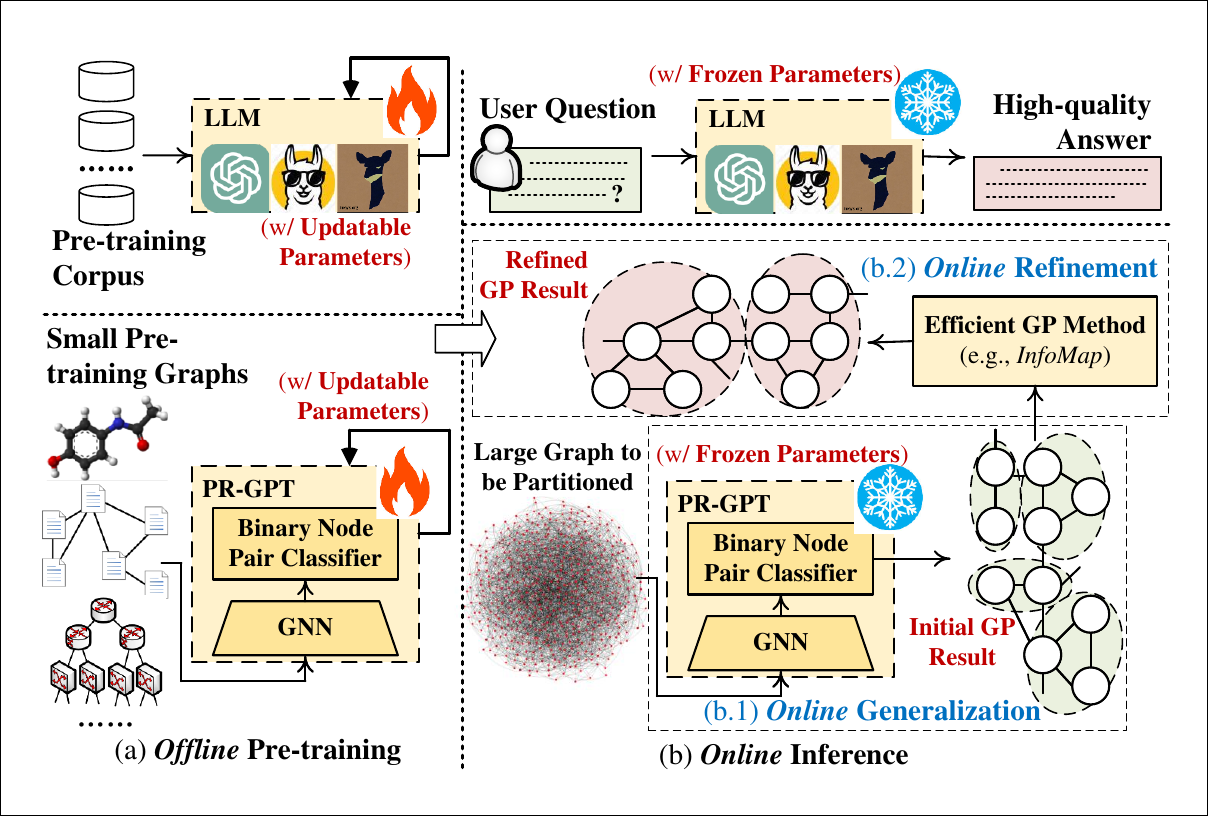}
  \vspace{-0.3cm}
  \caption{An overview about the applications of (\romannumeral1) foundation models (e.g., LLMs) and (\romannumeral2) our PR-GPT method, including the (a) \textit{offline pre-training} and (b) \textit{online inference}. The inference of PR-GPT includes the (b.1) \textit{online generalization} and (b.2) \textit{online refinement}.
  }\label{Fig:Overview}
  \vspace{-0.65cm}
\end{figure}

In this study, we explore the potential of deep graph learning (DGL) to obtain a better trade-off between the quality and efficiency of GP. Inspired by recent advances in foundation models (e.g., LLMs \cite{zhao2023survey}) and pre-training techniques \cite{cao2023pre,liu2023graphprompt}, we propose PR-GPT (\underline{P}re-trained \& \underline{R}efined \underline{G}raph \underline{P}ar\underline{T}itioning), a modification of our prior method \cite{qin2024pre}, with an overview shown in Fig.~\ref{Fig:Overview}.
It follows a pre-training \& refinement paradigm including the (\romannumeral1) \textit{offline pre-training}, (\romannumeral2) \textit{online generalization}, and (\romannumeral3) \textit{online refinement}.

Assume that one has enough time to prepare a well-trained DGL model in an offline way. We first pre-train the PR-GPT model (i.e., \textit{offline pre-training}) on a set of small graphs $\{ G_t \}$ (e.g., less than $5$K nodes) that cover various topology properties (e.g., node degrees and block sizes).
After that, we directly generalize the pre-trained model (with frozen model parameters) to large graphs $\{ G' \}$ (e.g., more than $1$M nodes) via inductive inference \cite{qin2023towards} and derive feasible GP results $\{ C' \}$ without re-training (i.e., \textit{online generalization}).
In existing pre-training techniques \cite{hu2019strategies,cao2023pre}, the online generalization after pre-training usually provides an initialization of model parameters, which are further fine-tuned w.r.t. different tasks.
Inspired by this motivation, we treat $C'$ as a good initialization of an efficient GP method (e.g., \textit{InfoMap} \cite{rosvall2008maps}) and adopt its output $\bar C'$ as a refined version of $C'$ (i.e., \textit{online refinement}).

Note that the application of PR-GPT is analogous to that of LLMs. For instance, users benefit from the online inference of ChatGPT, which can generate high-quality answers in just a few seconds but do not need to train an LLM from scratch using a great amount of resources. The \textit{online generalization} and \textit{refinement} of PR-GPT can benefit from inductive inference, which transfers the ability to derive high-quality GP results from pre-training data to new unseen graphs while ensuring high inference efficiency without re-training.
Experiments on the Graph Challenge benchmark demonstrate that PR-GPT can achieve faster GP without significant quality degradation, compared with running a refinement method from scratch.

The major contributions of this paper beyond our prior work \cite{qin2024pre} are summarized as follows.
\begin{itemize}
    \item We introduce PR-GPT, an advanced modification of our prior method, with fewer model parameters and better scalability. In contrast, directly applying our prior method to some large graphs in our experiments (e.g., more than $40$M edges) results in the out-of-memory exception.
    \item To the best of our knowledge, we are the first to submit a solution based on graph pre-training and inductive inference to the Graph Challenge benchmark, which involves the GP on synthetic graphs with various scales, topology properties, and ground-truth. Whereas, our prior method was only evaluated on static graphs without ground-truth.
    \item We also explore the ability of PR-GPT to support streaming GP while our prior work only considers static GP.
\end{itemize}

\section{Problem Statement \& Preliminaries}\label{Sec:Prob}
In general, a graph $G$ can be represented as a tuple $(V, E)$, where $V := \{ v_1, v_2, \cdots, v_N\}$ and $E := \{ (v_i, v_j) | v_i, v_j \in V \}$ are the sets of nodes and edges.
One can use an adjacency matrix ${\bf{A}} \in \{0, 1\}^{N \times N}$ to describe the topology structure of $G$, where ${\bf{A}}_{ij} = {\bf{A}}_{ji} = 1$ if $(v_i, v_j) \in E$ and ${\bf{A}}_{ij} = {\bf{A}}_{ji} = 0$ otherwise.
We adopt the problem statement of IEEE HPEC Graph Challenge \cite{uppal2018fast} and study the following GP problem.

\textbf{Graph Partitioning} (GP).
Given a graph $G$, (static) GP aims to partition the node set $V$ into $K$ disjoint subsets $C := (C_1, \cdots, C_K)$ (i.e., blocks or communities) s.t. (\romannumeral1) the linkage within each block is dense but (\romannumeral2) that between blocks is loose.
We consider the challenging $K$-agnostic GP, where the number of blocks $K$ is unknown. Namely, one should simultaneously determine $K$ and the corresponding block partition $C$.

\textbf{Streaming GP}. Graph Challenge provides two models to simulate streaming GP.
We consider the more challenging snowball model and leave the extension to emerging edges in future work. Given a graph $G$, the snowball model divides $V$ into $T$ disjoint subsets $(V_1, \cdots, V_T)$, with $V_t$ as the set of newly added nodes in the $t$-th step. Let ${\bar V}_t := \cup _{s = 1}^t{V_s}$ and ${\bar E}_t$ be the set of cumulative edges induced by ${\bar V}_t$. For each step $t$, streaming GP requires to derive a $K$-agnostic block partition based on the cumulative topology $({\bar V}_t, {\bar E}_t)$.

\textbf{Modularity Maximization}. GP can be formulated as the combinatorial optimization objective of modularity maximization \cite{newman2006modularity}. Given $G$ and $K$, it aims to obtain a partition $C$ that maximizes the modularity metric:
\begin{equation}\label{Eq:Mod}
    \mathop {\max }\limits_C ~ {\mathop{\rm Mod}\nolimits} (G,K): = \frac{1}{{2|E|}}\sum\limits_{r = 1}^K {\sum\limits_{{v_i},{v_j} \in {C_r}} {[{{\bf{A}}_{ij}} - \frac{{{d_i}{d_j}}}{{2|E|}}]} },
\end{equation}
where ${d_i}: = \sum\nolimits_i {{{\bf{A}}_{ij}}}$ is the degree of node $v_i$; $|E|$ is the number of edges. One can rewrite (\ref{Eq:Mod}) into the matrix form:
\begin{equation}\label{Eq:Mod-Mat}
    \mathop {\min }\limits_{\bf{H}}  - {\mathop{\rm tr}\nolimits} ({{\bf{H}}^T}{\bf{QH}})~{\rm{s.t.}}~{{\bf{H}}_{ir}} = \left\{ {\begin{array}{*{20}{l}}
    {1,{\rm{       }}{v_i} \in {C_r}}\\
    {0,{\rm{ otherwise}}}
    \end{array}} \right.,
\end{equation}
where ${\bf{Q}} \in {\mathbb{R}} ^ {N \times N}$ is defined as the modularity matrix with ${\bf{Q}}_{ij} := [{\bf{A}}_{ij} - d_i d_j / (2|E|)]$; ${\bf{H}} \in \{ 0, 1\}^{N \times K}$ indicates the block membership $C$.

\textbf{Pre-training \& Refinement}. As shown in Fig.~\ref{Fig:Overview}, the pre-training \& refinement paradigm includes the (\romannumeral1) \textit{offline pre-training}, (\romannumeral2) \textit{online generalization}, and (\romannumeral3) \textit{online refinement}.
For simplicity, we denote a DGL model as $C = f(G; \theta)$, which derives a block partition $C$ given a graph $G$, with $\theta$ as the set of model parameters to be learned.

In \textit{offline pre-training}, we generate a set of small graphs $\Gamma = \{ G_1, \cdots, G_M \}$ (e.g., less than $5$K nodes) using the generator of Graph Challenge. The generation of each $G_s \in \Gamma$ includes its topology $(V^{(s)}, E^{(s)})$ and corresponding block partition $C^{(s)}$. We then pre-train $f$ (e.g., iteratively updating $\theta$) based on $\{ (V^{(s)}, E^{(s)}) \}$ and $\{ C^{(s)} \}$ in an offline way.

After that, we generalize $f$ to new large graphs $\{ G' \}$ (e.g., more than $1$M nodes) with frozen $\theta$ in \textit{online generalization}. Based on the inductive inference of $f$, one can directly derive a feasible block partition $C'$ for $G'$ (e.g., via only one feed-forward propagation (FFP) of $f$) without re-training.
Inspired by existing pre-training and fine-tuning paradigm, we introduce \textit{online refinement}, where the derived partition $C'$ is used as a good initialization of an efficient $K$-agnostic GP method (e.g., \textit{InfoMap}) to further refine the quality of $C'$.

\textbf{Evaluation Protocol}.
Our evaluation is consistent with the application of foundation models as illustrated in Fig.~\ref{Fig:Overview}. Concretely, one can get high-quality answers from an LLM in just a few seconds during its \textit{online inference} phase and does not need to train it from scratch.
Assume that we have enough time to prepare a well-trained $f$ during the \textit{offline pre-training}, which is usually a one-time effort. Our evaluation focuses on the \textit{online generalization} and \textit{refinement} w.r.t. the GP on large graphs $\{ G' \}$. In contrast, we have to run most existing methods on $\{ G' \}$ from scratch, as they are inapplicable to inductive graph inference and thus cannot benefit from pre-training \cite{qin2023towards}.

\section{Methodology}\label{Sec:Meth}

\begin{figure}[]
  \centering
  \includegraphics[width=\linewidth, trim=18 18 18 18,clip]{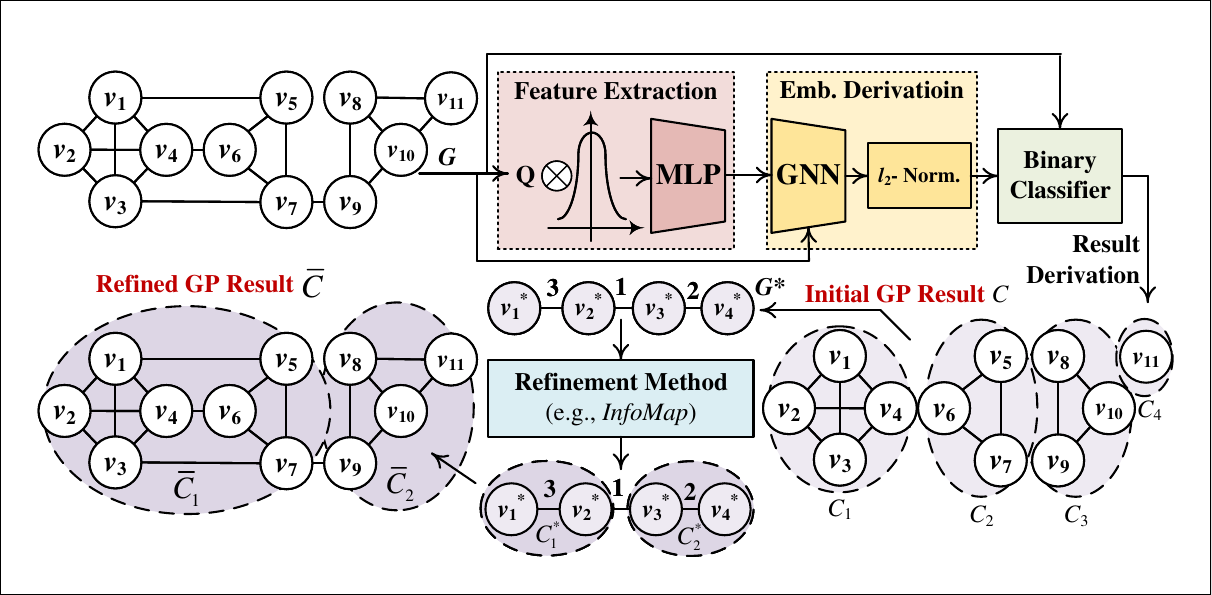}
  \vspace{-0.3cm}
  \caption{Model architecture and inference procedure of PR-GPT.
  }\label{Fig:Model}
  \vspace{-0.65cm}
\end{figure}

In this section, we detail our PR-GPT method. Fig.~\ref{Fig:Model} gives an overview of the model architecture and inference procedure.

\subsection{Model Architecture}
Similar to our prior method \cite{qin2024pre}, PR-GPT reformulates GP as the binary node pair classification and follows a GNN-based end-to-end architecture.
An auxiliary variable ${\bf{S}} \in \{ 0, 1 \}^{N \times N}$ is introduced to represent the binary classification result, where ${\bf{S}}_{ij} = {\bf{S}}_{ji} = 1$ if nodes $(v_i, v_j)$ are in the same block and ${\bf{S}}_{ij} = {\bf{S}}_{ji} = 0$ otherwise. The inference of PR-GPT only considers $\{ {\bf{S}}_{ij} \}$ w.r.t a small set of node pairs $P = \{ (v_i, v_j) \}$ ($|P| \ll N^2$), which are rearranged as a vector ${\bf{y}} \in \{ 0, 1\}^{|P|}$. We let ${\bf{y}}_s = {\bf{S}}_{ij} = {\bf{S}}_{ji}$ for each $p_s = (v_i, v_j) \in P$.

\subsubsection{\textbf{Feature Extraction}}
Let $k$ be the dimensionality of node embedding or features with $k \ll N$.
PR-GPT first uses the following feature extraction module to extract community-preserving features, arranged as a matrix ${\bf{\tilde X}} \in \mathbb{R}^{N \times k}$, from the modularity maximization objective (\ref{Eq:Mod-Mat}):
\begin{equation}\label{Eq:Feat-Ext}
    {\bf{\tilde X}} = {\mathop{\rm MLP}\nolimits} ({\bf{X}}) {\rm{~and~}}
    {\bf{X}} = {\bf{Q\Omega }} = {\bf{A\Omega }} - {\bf{d}}({{\bf{d}}^T}{\bf{\Omega }}),
\end{equation}
where ${\bf{\Omega}} \in \mathbb{R}^{N \times k}$ is a random matrix with ${\bf{\Omega}}_{ir} \sim \mathcal{N} (0, 1/k)$; ${\bf{d}} := [d_1, \cdots, d_N]^T \in \mathbb{Z}^{N}$ is a vector about node degrees.
It applies the Gaussian random projection \cite{arriaga2006algorithmic}, an efficient dimension reduction technique with rigorous bounds w.r.t. information loss, to the modularity matrix ${\bf{Q}}$ in (\ref{Eq:Mod-Mat}) that encodes key characteristics regarding implicit community structures.

Note that ${\bf{Q}} \in \mathbb{R}^{N \times N}$ is usually a dense matrix. Directly applying the random projection causes a complexity of $O(N^2 k)$ intractable for large graphs.
To reduce the complexity, our prior method introduces a sparsified matrix ${\bf{\tilde Q}} \in \mathbb{R}^{N \times N}$, where ${\bf{\tilde Q}}_{ij} = {\bf{Q}}$ if $(v_i, v_j) \in E$ and ${\bf{\tilde Q}}_{ij} = 0$ otherwise, and applies the random projection to ${\bf{\tilde Q}}$.
However, ${\bf{\tilde Q}}$ may lose some information encoded in ${\bf{Q}}$.
Instead of using ${\bf{\tilde Q}}$, PR-GPT still applies random projection to ${\bf{Q}}$ but follows the multiplication order described in (\ref{Eq:Feat-Ext}), which can reduce the complexity from $O(N^2k)$ to $O(|E|k + Nk)$.

We also apply a multi-layer perceptron (MLP) to the reduced features ${\bf{X}} \in \mathbb{R}^{N \times k}$, which leverages additional non-linearity for feature extraction.

\subsubsection{\textbf{Embedding Derivation}}
The extracted features ${\bf{\tilde Z}}$ are fed into a multi-layer GNN, which further derives community-preserving embeddings.
Inspired by SGC \cite{wu2019simplifying}, we remove all the learnable model parameters and non-linearity of GCN \cite{kipf2016semi} in our prior method, which improves the scalability.
Assume that there are $L$ GNN layers. PR-GPT derives embedding representations, arranged as ${\bf{\tilde Z}} \in \mathbb{R}^{N \times k}$, via
\begin{equation}\label{Eq:GNN}
    {\bf{\tilde Z}} = {\mathop{\rm LN}\nolimits} ({\bf{Z}}) {\rm{~and~}} {\bf{Z}} = {{{\bf{\tilde A}}}^L}{\bf{\tilde X}},
\end{equation}
where ${\mathop{\rm LN}\nolimits} ({\bf{Z}})$ is the row-wise $l_2$-normalization of the embedding matrix ${\bf{Z}} \in \mathbb{R}^{N \times k}$ (i.e., ${{\bf{Z}}_{i,:}} \leftarrow {{\bf{Z}}_{i,:}}/|{{\bf{Z}}_{i,:}}{|_2}$); ${\bf{\tilde A}}: = {{{\bf{\hat D}}}^{ - 1/2}}{\bf{\hat A}}{{{\bf{\hat D}}}^{ - 1/2}}$; ${\bf{\hat A}}: = {\bf{A}} + {{\bf{I}}_N}$ is the adjacency matrix with self-edges; ${\bf{\hat D}}$ is the degree diagonal matrix of ${\bf{\hat A}}$.

One can formulate the $l$-th GNN layer as ${\bf{Z}}^{(l)} = {\bf{\tilde A}} {\bf{Z}}^{(l-1)}$, with ${\bf{Z}}^{(0)} := {\bf{\tilde X}}$. ${\bf{Z}}^{(l)}_{i,:}$ is the intermediate representation of node $v_i$. It is also the weighted mean over features w.r.t. $\{ v_i \}  \cup n (v_i)$, with $n (v_i)$ as the neighbor set of $v_i$. This mechanism further enhances the ability of $\{ {\bf{Z}}^{(l)} \}$ to capture community structures, since it forces nodes $(v_i, v_j)$ with similar neighbors $(n(v_i), n(v_j))$ (i.e., dense local linkage) to have similar representations $({\bf{Z}}_{i, :}^{(l)}, {\bf{Z}}_{j,:}^{(l)})$.

Let ${\bf{\tilde z}}_i := {\bf{\tilde Z}}_{i,:}$ be the embedding of $v_i$. The $l_2$-normalization ensures that $|{\bf{\tilde z}}_i| = 1$ and thus $|{{{\bf{\tilde z}}}_i} - {{\bf{\tilde z}}_j}{|^2}  = 2 - 2{{\bf{z}}_i}{\bf{z}}_j^T$.

\subsubsection{\textbf{Binary Node Pair Classification}}
Given a node pair $(v_i, v_j)$, PR-GPT adopts the following binary classifier same as our prior method to estimate the classification result ${\bf{S}}_{ij}$ using corresponding embeddings $({\bf{\tilde z}}_i, {\bf{\tilde z}}_j)$:
\begin{equation}\label{Eq:Bin-Clf}
\begin{array}{l}
    {{{\bf{\hat S}}}_{ij}} = \exp ( - |{{{\bf{\tilde z}}}_i} - {{{\bf{\tilde z}}}_j}{|^2} {\tau _{ij}}) = \exp (2{\tau _{ij}} \cdot ({{{\bf{\tilde z}}}_i}{\bf{\tilde z}}_j^T - 1)),\\
    {\rm{with~}}{\tau _{ij}} = {g_s}({{{\bf{\tilde z}}}_i}){g_d}({{{\bf{\tilde z}}}_j}),
\end{array}
\end{equation}
where ${\bf{\hat S}}_{ij} \in [0, 1]$ denotes the estimation of ${\bf{S}}_{ij}$; $g_s$ and $g_d$ are two MLPs with the same configurations; $\tau_{ij}$ is a pair-wise parameter determined by $({\bf{\tilde z}}_i, {\bf{\tilde z}}_j)$. Namely, $\{ {\bf{S}}_{ij} \}$ is estimated via a combination of the (\romannumeral1) Euclidean distance and (\romannumeral2) inner product w.r.t. corresponding embeddings $\{ ({\bf{\tilde z}}_i, {\bf{\tilde z}}_j) \}$.

\begin{algorithm}[t]\small
\caption{\small
Result Derivation Given a Graph
}
\label{Alg:Res}
\LinesNumbered
\KwIn{input graph $G = (V, E)$}
\KwOut{a feasible GP result $C$ w.r.t. $G$}
derive $\bf{\hat y}$ w.r.t. $E$ via one FFP of the model\\
${\tilde E} \leftarrow \emptyset$ //Initialize edge set of auxiliary graph ${\tilde G}$\\
\For{{\bf{each}} {\rm{node pair (i.e., edge)}} $e_s = (v_i, v_j) \in E$}
{
    \If{${\bf{\hat y}}_s > 0.5$}
    {
        add $e_s = (v_i, v_j)$ to ${\tilde E}$
    }
}
extract connected components of ${\tilde G}$ via DFS/BFS on ${\tilde E}$\\
treat each component as a block to form $C$
\end{algorithm}

\subsubsection{\textbf{Result Derivation}}

Given a graph $G = (V, E)$, the overall procedure to derive a feasible GP result is summarized in Algorithm~\ref{Alg:Res}. Fig.~\ref{Fig:Model} also gives a running example.

In line 1, we derive $\{ {\bf{\hat S}}_{ij}\}$ for all the node pairs (i.e., edges) in the edge set $E$ of $G$ and rearrange them as a vector ${\bf{\hat y}}$.
In lines 2-5, an auxiliary graph ${\tilde G} = (V, {\tilde E})$ is constructed based on ${\bf{\hat y}}$. ${\tilde G}$ shares the same node set $V$ as $G$ but has a different edge set ${\tilde E}$. Concretely, we add $e_s = (v_i, v_j)$ to ${\tilde E}$ if ${\bf{\hat y}}_s > 0.5$.
After that, the constructed ${\tilde G}$ may contain multiple connected components (e.g., $\{ v_1, v_2, v_3, v_4\}$ and $\{ v_5, v_6, v_7 \}$ in Fig.~\ref{Fig:Model}). In particular, edges $\{ e_s \}$ in the same component are with high values $\{ {\bf{\hat y}}_s \}$. It indicates that the associated nodes are more likely to be partitioned into the same block.
In lines 6-7, we extract all the connected components of $\tilde G$ via the DFS/BFS on ${\tilde E}$ and treat each component as a feasible block.

In addition to $E$, our prior method also sampled other node pairs (e.g., $(v_5, v_8)$ in Fig.~\ref{Fig:Model}) when constructing $\tilde G$. PR-GPT removes this sampling procedure to further improve efficiency.

Our experiments demonstrate that even with the aforementioned simplifications, PR-GPT can still achieve impressive GP quality on the Graph Challenge benchmark.

\subsection{Offline Pre-training}
PR-GPT adopts the same setup of \textit{offline pre-training} as our prior method. We first generate a set of small pre-training graphs $\Gamma = \{ G_1, \cdots, G_M\}$ using the standard generator of Graph Challenge. Instead of fixing generator parameters, we simulate various properties (e.g., distributions of degrees and block sizes) by sampling these parameters from certain distributions, which can increase the diversity of pre-training data.
The pre-training of PR-GPT combines the (\romannumeral1) unsupervised modularity maximization and (\romannumeral2) supervised binary cross-entropy objectives for each graph $G_t \in \Gamma$.
Due to space limits, we omit details of pre-training data generation and pre-training algorithm, which can be found in our prior work \cite{qin2024pre} and code.

\begin{algorithm}[t]\small
\caption{\small
\textit{Online Generalization} and \textit{Refinement}
}
\label{Alg:Inf}
\LinesNumbered
\KwIn{a large graph $G' = (V', E')$ to be partitioned}
\KwOut{a refined GP result $\bar C'$ w.r.t. $G'$}
get input features ${\bf{\tilde X'}}$ w.r.t. $\mathcal{G}'$ via (\ref{Eq:Feat-Ext})\\
get initial GP result $C'$ w.r.t. $G'$ via Algorithm~\ref{Alg:Res}\\
construct weighted super-graph $G^*$ based on $C'$\\
get refined $\bar C'$ via a refinement method w/ $G^*$ as input
\end{algorithm}

\subsection{Online Inference for Static GP}

As described in Algorithm~\ref{Alg:Inf}, the \textit{online inference} of PR-GPT includes the \textit{online generalization} (i.e., lines 1-2) and \textit{online refinement} (i.e., lines 3-4).

\subsubsection{\textbf{Online Generalization}}
After the \textit{offline pre-training}, we can generalize PR-GPT to a large graph $G'$ with frozen parameters $\theta$ and derive a feasible partition $C'$ via Algorithm~\ref{Alg:Res}.

\subsubsection{\textbf{Online Refinement}}
We further treat the derived $C'$ as a good initialization of an existing $K$-agnostic GP method and apply this method to refine $C'$.
Concretely, the initialization can be in the form of a weighted super-graph $G^*$ w.r.t. $C'$, where we merge nodes in each block as a super-node (e.g., $v_1^* = C_1 = \{ v_1, v_2, v_3, v_4\}$ in Fig.~\ref{Fig:Model}) and set the number of between-block edges as the weight of corresponding super-edge (e.g., $3$ for $(v_1^*, v_2^*)$ in Fig.~\ref{Fig:Model}). We use $G^*$ as the input of a $K$-agnostic GP method that can handle weighted graphs (e.g., \textit{InfoMap} \cite{rosvall2008maps} and \textit{Locale} \cite{wang2020community} in our experiments) and derive a GP result $C^*$ w.r.t. $G^*$. $C^*$ is then recovered to a partition $\bar C'$ w.r.t. $G'$, which is a refined version of $C'$.

Compared with running a refinement method on $G'$ from scratch, \textit{online refinement} may be much more efficient, since it reduces the number of nodes to be processed (e.g., reducing $11$ nodes to $4$ super-nodes in Fig.~\ref{Fig:Model}). Therefore, PR-GPT has the potential to achieve faster GP w.r.t. the refinement method.

\subsection{Extension to Streaming GP}\label{Sec:Stream}
The \textit{online generalization} and \textit{refinement} of PR-GPT shares a motivation similar to that of some streaming GP approaches \cite{zhuzhunashvili2017preconditioned,uppal2018fast}, where the partition of current step provides an \textbf{initialization} for next step. In each step, these streaming approaches incrementally update their GP results to \textbf{avoid running the base algorithm from scratch}. Similarly, PR-GPT tries to achieve faster GP compared with a refinement method by reducing the number of nodes to be processed.
Due to these similar motivations, we believe that PR-GPT has the potential to support streaming GP.

As stated in Section~\ref{Sec:Prob}, we consider the snowball model, where new nodes are added in each step. The inductive inference of DGL allows RP-GPT to directly generalize the pre-trained model parameters to new topology of each step without re-training. Our analysis about inference time (see Table~\ref{Tab:Time}) shows that \textit{online refinement} is the major bottleneck of PR-GPT.
We adopt a naive strategy to directly run the non-bottleneck \textit{online generalization} procedure from scratch in each step.
Our experiments demonstrate that even without incremental updating for \textit{online generalization}, PR-GPT can still obtain efficiency improvement for streaming GP, which is consistent with some previous work \cite{uppal2018fast}, based on its mechanism of reducing the scale of a graph to be processed.

\subsection{Complexity Analysis}
To derive embeddings $\{ {\bf{\tilde z}}_i \}$, the complexities of (\romannumeral1) feature extraction described in (\ref{Eq:Feat-Ext}) and (\romannumeral2) one FFP of GNN defined in (\ref{Eq:GNN}) are $O(|E|k + Nk)$ and $O(|E|Lk + Nk)$, with $L$ as the number of GNN layers.
The complexities of (\romannumeral3) one FFP of binary classifier (\ref{Eq:Bin-Clf}) to construct ${\tilde G}$ and (\romannumeral4) extracting connected components of ${\tilde G}$ via DFS/BFS are $O(|E|k^2)$ and $O(N + |\tilde E|)$, where $|\tilde E|$ is the number of edges in ${\tilde G}$, with $|\tilde E| \le |E|$.
Therefore, the overall complexity of \textit{online generalization} is no more than $O(|E|k(L+k) + Nk)$, with $k, L \ll N, |E|$.

The complexity of \textit{online refinement} depends on the concrete refinement method (e.g., \textit{InfoMap}), which is usually efficient. Note that any improvement regarding the efficiency of a refinement method (e.g., better parallel implementations) can also further improve the efficiency of PR-GPT.

\section{Experiments}\label{Sec:Exp}

\subsection{Experiment Setups}\label{Sec:Exp-Set}

\begin{table}[t]
\caption{Detailed Statistics of the Generated Benchmark Datasets}\label{Tab:Data}
\centering
\begin{tabular}{l|l|l|l|l}
\hline
$N$ & $|E|$ & $K$ & \textbf{Degrees} & \textbf{Density}  \\ \hline
10K & 402K-449K & 25 & 6-194 & $8 \times 10^{-3}$  \\
50K & 2.02M-2.03M & 44 & 8-246 & $2 \times 10^{-3}$ \\
100K & 4.05M-4.07M & 56 & 6-242 & $8 \times 8^{-4}$ \\
500K & 20.3M-20.3M & 98 & 4-230 & $1 \times 10^{-4}$ \\
1M & 40.6M-40.7M & 125 & 4-264 & $8 \times 10^{-5}$ \\ \hline
\end{tabular}
\vspace{-0.2cm}
\end{table}

We followed standard setups of the IEEE HPEC Graph Challenge benchmark \cite{kao2017streaming} to validate the effectiveness of PR-GPT for both static and streaming GP.

\subsubsection{\textbf{Datasets}}
We considered the hardest setting with the (\romannumeral1) ratio between the number of within-block edges and between edges and (\romannumeral2) block size heterogeneity set to $2.5$ and $3$. The standard generator of Graph Challenge was used to generate graphs with different scales, where we respectively set the number of nodes $N$ to $10$K, $50$K, $100$K, $500$K, and $1$M. For each setting, we independently generated five graphs and reported corresponding average evaluation results. Table~\ref{Tab:Data} summarizes statistics of the generated datasets, where $|E|$ and $K$ denote the numbers of edges and blocks.

\subsubsection{\textbf{Baselines}}
We compared PR-GPT with seven baselines that can tackle $K$-agnostic GP, including (\romannumeral1) \textit{MC-SBM} \cite{peixoto2014efficient}, (\romannumeral2) \textit{Par-SBM} \cite{peng2015scalable}, (\romannumeral3) \textit{Louvain} \cite{blondel2008fast}, (\romannumeral4) \textit{RaftGP-C} \cite{gao2023raftgp}, (\romannumeral5) \textit{RaftGP-M} \cite{gao2023raftgp}, (\romannumeral6) \textit{InfoMap} \cite{rosvall2008maps}, and (\romannumeral7) \textit{Locale} \cite{wang2020community}.

In particular, \textit{MC-SBM} is the standard baseline of Graph Challenge. \textit{Locale} is an advanced modification of \textit{Louvain}. \textit{RaftGP-C} and \textit{RaftGP-M} are two variants of the innovation award winner of Graph Challenge 2023, which are GNN-based embedding approaches without (pre-)training. We adopt \textit{InfoMap} and \textit{Locale} as two example refinement methods of PR-GPT and highlight the improvement in quality and efficiency w.r.t. running these refinement methods from scratch. Note that PR-GPT can also be easily extended to leverage other efficient GP approaches (e.g., \textit{Louvain}) for \textit{online refinement}.

\subsubsection{\textbf{Evaluation Metrics}}
We followed the evaluation criteria of Graph Challenge to adopt \textit{accuracy} (AC), \textit{adjusted random index} (ARI), \textit{precision}, and \textit{recall} as quality metrics. Given precision and recall, we also reported the corresponding \textit{F1-score}. The \textit{inference time} (sec) of a method was adopted as the efficiency metric, based on the evaluation protocol described in Section~\ref{Sec:Prob}. We defined that a method encounters the out-of-time exception if it cannot derive a feasible GP result within $10,000$ seconds.

\subsubsection{\textbf{Parameter} \& \textbf{Environment Settings}}
Let $k$ be the feature or embedding dimensionality. $L_{\rm{F}}$, $L_{\rm{GNN}}$, and $L_{\rm{BC}}$ denote the numbers of MLP layers in (\ref{Eq:Feat-Ext}), GNN layers in (\ref{Eq:GNN}), and MLP layers in (\ref{Eq:Bin-Clf}), respectively. We set $(k, L_{\rm{F}}, L_{\rm{GNN}}, L_{\rm{BC}}) = (32, 2, 2, 4)$ for PR-GPT on all the datasets.
Other parameter settings of PR-GTP can be checked in our code.

We used Python 3.8 to implement PR-GPT, where the model architecture and Algorithm~\ref{Alg:Res} were implemented via PyTorch 1.10 and SciPy 1.10 (with BFS/DFS supported by the efficient `scipy.sparse.csgraph.connected\_components' function).
Moreover, we adopted the official or widely-used (C/C++ or Python) implementations of all the baselines and tuned their parameters to report the best quality.

All the experiments were conducted on a server with one Intel 14-core CPU, one 24GB memory GPU, 45GB main memory, and Ubuntu 20.04 OS.

\subsection{Evaluation of Static GP}

\begin{table}[t]\scriptsize
\caption{Evaluation Results of Static GP with $N$=10K}\label{Tab:Static-GP-10K}
\centering
\begin{tabular}{l|l|lll}
\hline
\textbf{Methods} & \textbf{Time}$\downarrow${\tiny (s)} & \textbf{AC}$\uparrow${\tiny (\%)} & \textbf{ARI}$\uparrow${\tiny (\%)} & \textbf{F1}$\uparrow$(RCL, PCN){\tiny (\%)}\\ \hline
MC-SBM & 217.06 & 99.32 & 99.51 & 99.53 (99.17, 99.91) \\
Par-SBM & 2.34 & 99.89 & 99.97 & 99.97 (99.97, 99.98) \\
Louvain & 3.99 & 94.68 & 95.71 & 95.94 (99.90, 92.38) \\
RaftGP-C & 11.01 & 99.32 & 99.15 & 99.20 (99.86, 98.56) \\
RaftGP-M & 10.34 & 99.27 & 99.09 & 99.14 (99.86, 98.45) \\ \hline
\textit{InfoMap} & 1.18 & 99.89 & 99.96 & 99.96 (99.97, 99.96) \\
PR-GPT{\tiny (\textbf{IM})} & \textbf{0.80} & \textbf{99.94} & 99.96 & 99.96 (99.98, 99.95) \\
~~\textbf{Improv.} & +32.2\% & +0.05\% & +0.0\% & +0.0\% \\ \hline
\textit{Locale} & 3.20 & 95.15 & 97.18 & 97.33 (99.96, 94.86) \\
PR-GPT{\tiny (\textbf{Lcl})} & \textbf{1.56} & \textbf{96.81} & \textbf{98.08} & \textbf{98.19} (99.94, 96.50) \\
~~\textbf{Improv.} & +51.3\% & +1.7\% & +0.9\% & +0.9\% \\ \hline
\end{tabular}
\end{table}

\begin{table}[t]\scriptsize
\caption{Evaluation Results of Static GP with $N$=50K}\label{Tab:Static-GP-50K}
\centering
\begin{tabular}{l|l|lll}
\hline
\textbf{Methods} & \textbf{Time}$\downarrow${\tiny (s)} & \textbf{AC}$\uparrow${\tiny (\%)} & \textbf{ARI}$\uparrow${\tiny (\%)} & \textbf{F1}$\uparrow$(RCL, PCN){\tiny (\%)}\\ \hline
MC-SBM & 2553.67 & 99.33 & 99.19 & 99.22 (98.51, 99.97) \\
Par-SBM & 19.77 & 99.00 & 99.34 & 99.36 (99.98, 98.76) \\
Louvain & 29.13 & 83.60 & 72.43 & 73.51 (99.91, 59.00) \\
RaftGP-C & 63.91 & 99.18 & 98.52 & 98.57 (99.91, 97.29) \\
RaftGP-M & 63.16 & 99.00 & 98.16 & 98.22 (99.89, 96.66) \\ \hline
\textit{InfoMap} & 11.67 & 99.88 & 99.96 & 99.96 (99.97, 99.94) \\
PR-GPT{\tiny (\textbf{IM})} & \textbf{6.61} & 99.63 & 99.63 & 99.64 (99.94, 99.35) \\
~~\textbf{Improv.} & +43.6\% & -0.3\% & -0.3\% & -0.3\% \\ \hline
\textit{Locale} & 23.75 & 93.97 & 94.44 & 94.62 (99.96, 89.92) \\
PR-GPT{\tiny (\textbf{Lcl})} & \textbf{13.44} & \textbf{94.49} & 94.41 & 94.59 (99.90, 89.85) \\
~~\textbf{Improv.} & +43.4\% & +0.6\% & -0.03\% & -0.03\% \\ \hline
\end{tabular}
\end{table}

\begin{table}[t]\scriptsize
\caption{Evaluation Results of Static GP with $N$=100K}\label{Tab:Static-GP-100K}
\centering
\begin{tabular}{l|l|lll}
\hline
\textbf{Methods} & \textbf{Time}$\downarrow${\tiny (s)} & \textbf{AC}$\uparrow${\tiny (\%)} & \textbf{ARI}$\uparrow${\tiny (\%)} & \textbf{F1}$\uparrow$(RCL, PCN){\tiny (\%)}\\ \hline
MC-SBM & 9240.27 & 99.00 & 98.93 & 98.95 (97.98, 99.99) \\
Par-SBM & 52.68 & 98.97 & 99.62 & 99.63 (99.99, 99.26) \\
Louvain & 68.03 & 74.08 & 59.99 & 61.28 (99.94, 44.40) \\
RaftGP-C & 138.58 & 99.49 & 99.04 & 99.06 (99.93, 98.23) \\
RaftGP-M & 133.14 & 99.75 & 99.58 & 99.59 (99.95, 99.24) \\ \hline
\textit{InfoMap} & 28.85 & 99.93 & 99.97 & 99.97 (99.97, 99.96) \\
PR-GPT{\tiny (\textbf{IM})} & \textbf{16.92} & 99.82 & 99.81 & 99.81 (99.85, 99.78) \\
~~\textbf{Improv.} & +41.4\% & -0.1\% & -0.2\% & -0.2\% \\ \hline
\textit{Locale} & 55.33 & 89.91 & 86.87 & 87.21 (99.97, 77.54) \\
PR-GPT{\tiny (\textbf{Lcl})} & \textbf{31.79} & \textbf{90.84} & \textbf{89.79} & \textbf{90.05} (99.83, 82.12) \\
~~\textbf{Improv.} & +42.5\% & +1.0\% & +3.4\% & +3.3\% \\ \hline
\end{tabular}
\end{table}

\begin{table}[t]\scriptsize
\caption{Evaluation Results of Static GP with $N$=500K}\label{Tab:Static-GP-500K}
\centering
\begin{tabular}{l|l|lll}
\hline
\textbf{Methods} & \textbf{Time}$\downarrow${\tiny (s)} & \textbf{AC}$\uparrow${\tiny (\%)} & \textbf{ARI}$\uparrow${\tiny (\%)} & \textbf{F1}$\uparrow$(RCL, PCN){\tiny (\%)}\\ \hline
MC-SBM & OOT & OOT & OOT & OOT \\
Par-SBM & 305.65 & 97.69 & 98.69 & 98.71 (99.99, 97.49) \\
Louvain & 561.62 & 39.75 & 20.85 & 22.73 (99.90, 12.85) \\
RaftGP-C & OOM & OOM & OOM & OOM \\
RaftGP-M & OOM & OOM & OOM & OOM \\ \hline
\textit{InfoMap} & 245.78 & 99.42 & 99.75 & 99.75 (99.76, 99.74) \\
PR-GPT{\tiny (\textbf{IM})} & \textbf{194.68} & 99.16 & 98.77 & 98.79 (99.76, 97.91) \\
~~\textbf{Improv.} & +20.8\% & -0.3\% & -1.0\% & -1.0\% \\ \hline
\textit{Locale} & 464.72 & 61.25 & 38.45 & 39.77 (99.96, 24.96) \\
PR-GPT{\tiny (\textbf{Lcl})} & \textbf{325.28} & \textbf{69.71} & \textbf{51.84} & \textbf{52.79} (99.79, 35.99) \\
~~\textbf{Improv.} & +30.0\% & +13.8\% & +34.8\% & +32.7\% \\ \hline
\end{tabular}
\end{table}

\begin{table}[t]\scriptsize
\caption{Evaluation Results of Static GP with $N$=1M}\label{Tab:Static-GP-1M}
\centering
\begin{tabular}{l|l|lll}
\hline
\textbf{Methods} & \textbf{Time}$\downarrow${\tiny (s)} & \textbf{AC}$\uparrow${\tiny (\%)} & \textbf{ARI}$\uparrow${\tiny (\%)} & \textbf{F1}$\uparrow$(RCL, PCN){\tiny (\%)}\\ \hline
MC-SBM & OOT & OOT & OOT & OOT \\
Par-SBM & 761.63 & 98.90 & 99.43 & 99.44 (99.99, 98.89) \\
Louvain & 1266.81 & 25.10 & 13.76 & 15.41 (99.87, 8.38) \\
RaftGP-C & OOM & OOM & OOM & OOM \\
RaftGP-M & OOM & OOM & OOM & OOM \\ \hline
\textit{InfoMap} & 680.75 & 98.64 & 99.35 & 99.36 (99.45, 99.26) \\
PR-GPT{\tiny (\textbf{IM})} & \textbf{530.42} & \textbf{99.07} & \textbf{99.48} & \textbf{99.49} (99.55, 99.42) \\
~~\textbf{Improv.} & +22.1\% & +0.4\% & +0.1\% & +0.1\% \\ \hline
\textit{Locale} & 1216.16 & 49.70 & 24.84 & 26.19 (99.95, 15.13) \\
PR-GPT{\tiny (\textbf{Lcl})} & \textbf{996.65} & \textbf{62.23} & \textbf{35.53} & \textbf{36.62} (99.81, 22.58) \\
~~\textbf{Improv.} & +18.1\% & +25.2\% & +43.0\% & +39.8\% \\ \hline
\end{tabular}
\vspace{-0.3cm}
\end{table}


\begin{table}[t]\scriptsize
\caption{Detailed Inference Time (sec) of PR-GPT}\label{Tab:Time}
\centering
\begin{tabular}{c|l|l|cccl}
\hline
\multicolumn{1}{l|}{\textit{N}} &  & \textbf{Total} & \multicolumn{1}{l}{\textbf{Feat}} & \multicolumn{1}{l}{\textbf{FFP}} & \multicolumn{1}{l}{\textbf{Init}} & \textbf{Refine} \\ \hline
\multirow{2}{*}{10K} & PR-GPT (IM) & 0.80 & \multirow{2}{*}{0.01} & \multirow{2}{*}{0.03} & \multirow{2}{*}{0.51} & 0.25 \\
 & PR-GPT (Lcl) & 1.56 &  &  &  & 1.00 \\ \hline
\multirow{2}{*}{50K} & PR-GPT (IM) & 6.61 & \multirow{2}{*}{0.04} & \multirow{2}{*}{0.16} & \multirow{2}{*}{2.91} & 3.49 \\
 & PR-GPT (Lcl) & 13.44 &  &  &  & 10.33 \\ \hline
\multirow{2}{*}{100K} & PR-GPT (IM) & 16.92 & \multirow{2}{*}{0.08} & \multirow{2}{*}{0.33} & \multirow{2}{*}{6.45} & 10.06 \\
 & PR-GPT (Lcl) & 31.79 &  &  &  & 24.93 \\ \hline
\multirow{2}{*}{500K} & PR-GPT (IM) & 194.68 & \multirow{2}{*}{0.41} & \multirow{2}{*}{1.69} & \multirow{2}{*}{46.64} & 145.94 \\
 & PR-GPT (Lcl) & 325.28 &  &  &  & 276.54 \\ \hline
\multirow{2}{*}{1M} & PR-GPT (IM) & 530.42 & \multirow{2}{*}{0.80} & \multirow{2}{*}{3.34} & \multirow{2}{*}{113.14} & 413.13 \\
 & PR-GPT (Lcl) & 996.65 &  &  &  & 879.36 \\ \hline
\end{tabular}
\end{table}

\begin{figure*}
\centering
 \begin{minipage}{0.118\linewidth}
 \subfigure[IM, Time$\downarrow$ (s)]{
  \includegraphics[width=\textwidth,trim=0 0 12 25,clip]{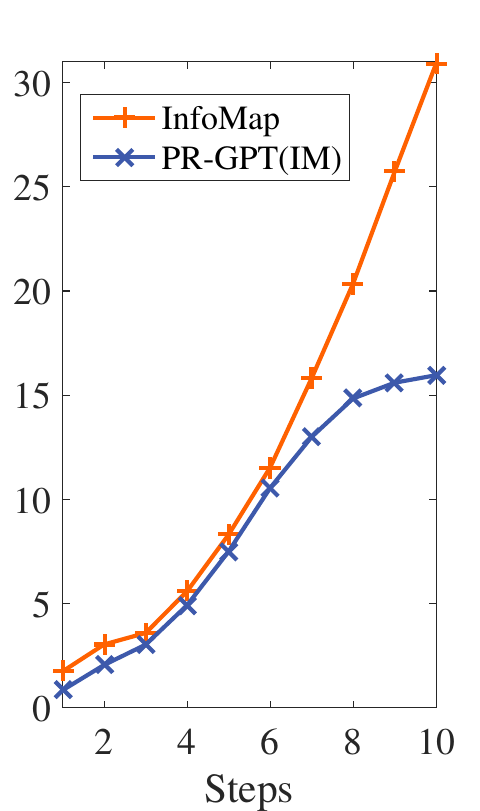}}
 \end{minipage}
 \begin{minipage}{0.118\linewidth}
 \subfigure[IM, AC$\uparrow$]{
  \includegraphics[width=\textwidth,trim=0 0 12 25,clip]{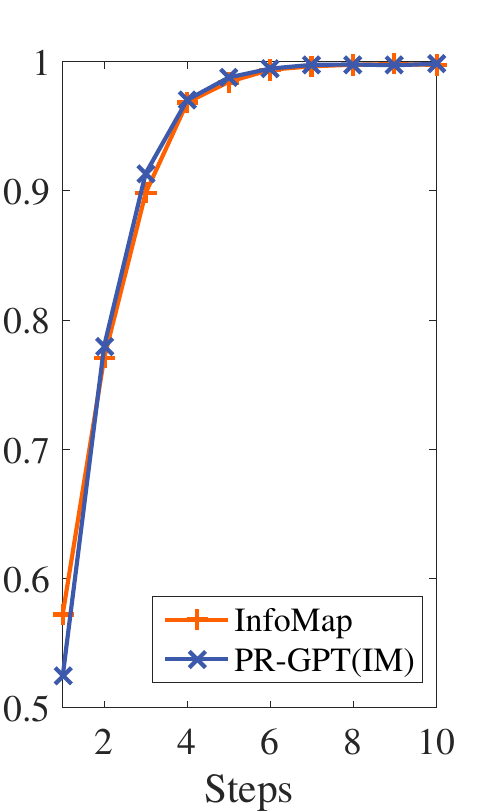}}
 \end{minipage}
 \begin{minipage}{0.118\linewidth}
 \subfigure[IM, ARI$\uparrow$]{
  \includegraphics[width=\textwidth,trim=0 0 12 25,clip]{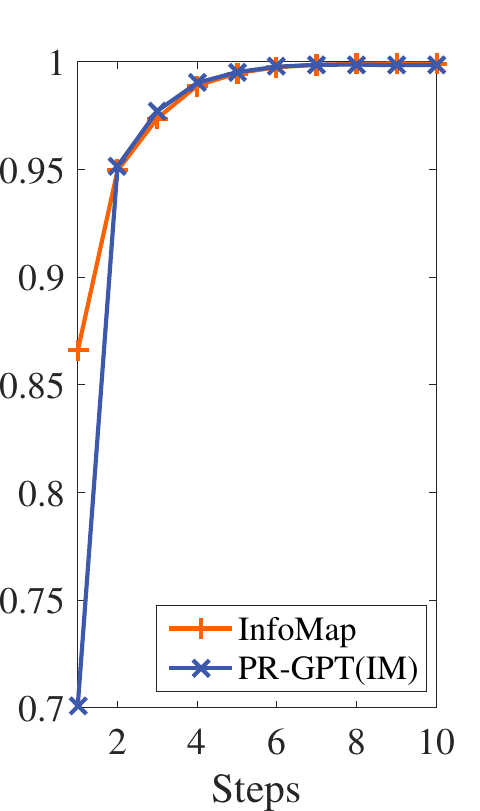}}
 \end{minipage}
 \begin{minipage}{0.118\linewidth}
 \subfigure[IM, F1$\uparrow$]{
  \includegraphics[width=\textwidth,trim=0 0 12 25,clip]{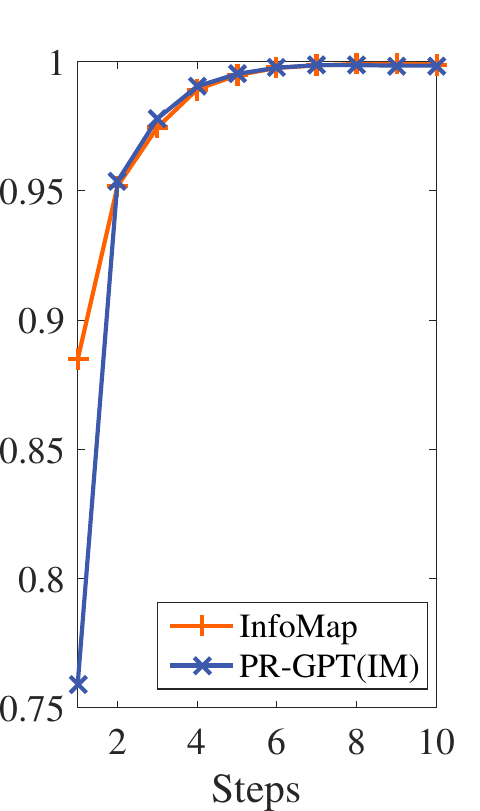}}
 \end{minipage}
 \begin{minipage}{0.118\linewidth}
 \subfigure[Lcl, Time$\downarrow$ (s)]{
  \includegraphics[width=\textwidth,trim=0 0 12 25,clip]{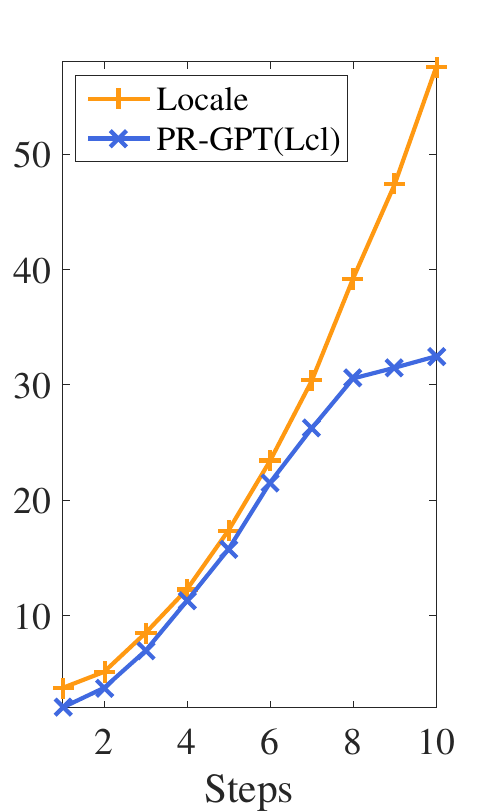}}
 \end{minipage}
 \begin{minipage}{0.118\linewidth}
 \subfigure[Lcl, AC$\uparrow$]{
  \includegraphics[width=\textwidth,trim=0 0 12 25,clip]{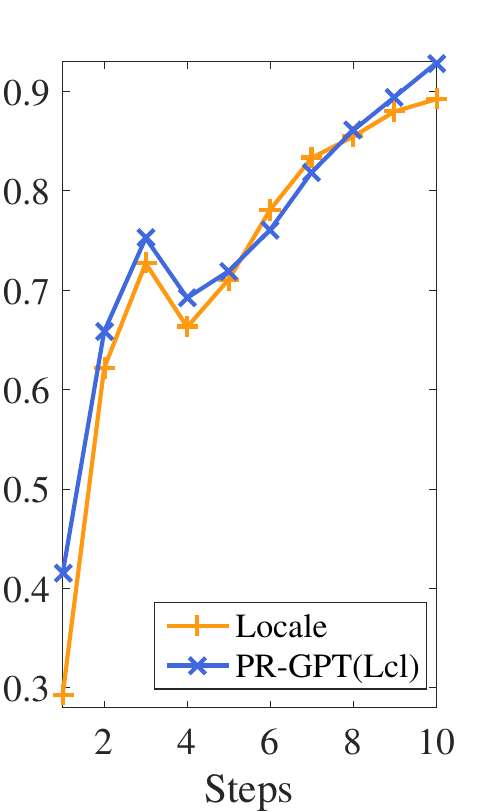}}
 \end{minipage}
 \begin{minipage}{0.118\linewidth}
 \subfigure[Lcl, ARI$\uparrow$]{
  \includegraphics[width=\textwidth,trim=0 0 12 25,clip]{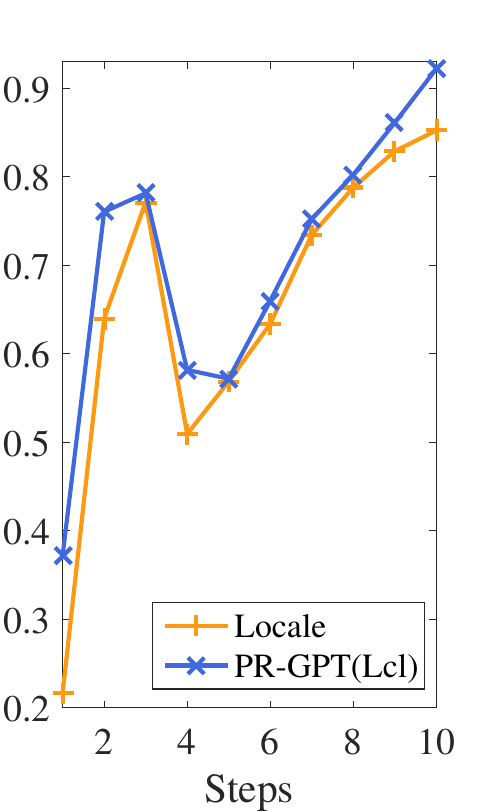}}
 \end{minipage}
 \begin{minipage}{0.118\linewidth}
 \subfigure[Lcl, F1$\uparrow$]{
  \includegraphics[width=\textwidth,trim=0 0 12 25,clip]{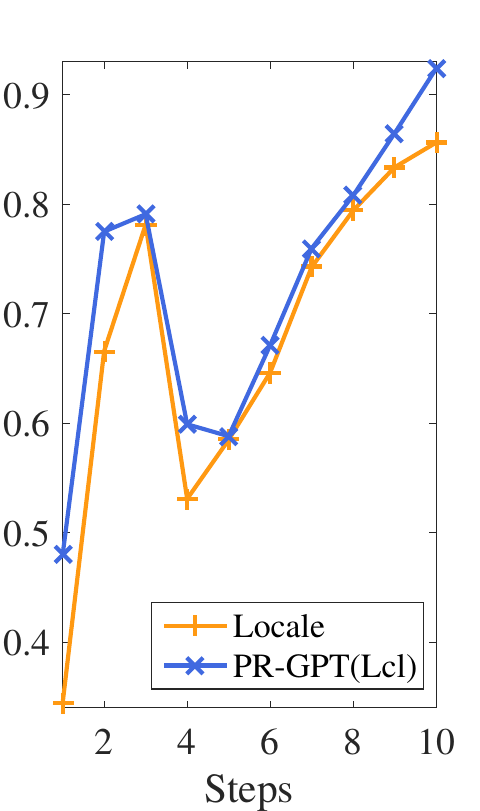}}
 \end{minipage}
\caption{Evaluation results of streaming GP.}\label{Fig:Stream-GP}
\vspace{-0.2cm}
\end{figure*}

The average evaluation results of static GP over five generated graphs w.r.t. each setting of the dataset are shown in Tables~\ref{Tab:Static-GP-10K}, \ref{Tab:Static-GP-50K}, \ref{Tab:Static-GP-100K}, \ref{Tab:Static-GP-500K}, and \ref{Tab:Static-GP-1M}, where metrics of PR-GPT are in \textbf{bold} if they achieve improvement w.r.t. the corresponding refinement methods; OOT and OOM represent the out-of-time and out-of-memory exceptions; `(IM)' and `(Lcl)' indicate that \textit{InfoMap} and \textit{Locale} are used as the refinement method.
We also report the corresponding runtime (sec) of each \textit{online inference} step of PR-GPT in Table~\ref{Tab:Time}, where `Feat', `FFP', `Init', and `Refine' represent the runtime of (\romannumeral1) feature extraction described in (\ref{Eq:Feat-Ext}), (\romannumeral2) one FFP of the model, (\romannumeral3) initial result derivation (i.e., Algorithm~\ref{Alg:Res}), and (\romannumeral4) \textit{online refinement}.

On all the datasets, two variants of PR-GPT achieve significant improvement of efficiency (e.g., more than 20\%) w.r.t. the corresponding refinement methods while the quality degradation is less than 1\%.
In summary, PR-GPT achieves the best efficiency and is always in the top groups with the best quality. It indicates that the pre-training \& refinement paradigm of PR-GPT can help ensure a better trade-off between the quality and efficiency of static GP.

Surprisingly, PR-GPT can even obtain improvement for both aspects in some cases. Compared with \textit{InfoMap}, \textit{Locale} is a weaker refinement method in terms of quality. PR-GPT can even ensure significant quality improvement (e.g., more than 10\%) when \textit{Locale} suffers from poor quality metrics (e.g., on datasets with $N=500$K and $1$M).
The aforementioned results demonstrate that the \textit{offline pre-training} (on historical small graphs) may also help resist noise and improve the GP quality (on new large graphs) compared with running a refinement method from scratch. In contrast, most existing GP approaches cannot benefit from pre-training and inductive inference.

According to Table~\ref{Tab:Time}, \textit{online refinement} is the major bottleneck of PR-GPT, depending on the concrete refinement method. In particular, the runtime of \textit{online refinement} is much smaller than that of running a refinement method from scratch. It verifies our motivation that the \textit{online generalization} can derive a good initialization (i.e., a weighted super-graph $G^*$) with a much smaller scale (e.g., in terms of the number of nodes to be processed) and thus help achieve faster GP.

\begin{table}[t]
\caption{Detailed Variation of $N$ and $\tilde N$ in Streaming GP}\label{Tab:Stream-N}
\centering
\begin{tabular}{l|p{0.2cm}p{0.2cm}p{0.2cm}p{0.2cm}p{0.2cm}p{0.2cm}p{0.2cm}p{0.2cm}p{0.2cm}l}
\hline
Steps & 1 & 2 & 3 & 4 & 5 & 6 & 7 & 8 & 9 & 10 \\ \hline
$N$ & 10K & 20K & 30K & 40K & 50K & 60K & 70K & 80K & 90K & 100K \\
$\tilde N$ & 8K & 18K & 27K & 37K & 46K & 54K & 60K & 63K & 63K & 60K \\ \hline
\end{tabular}
\vspace{-0.3cm}
\end{table}

\subsection{Evaluation of Streaming GP}

As discussed in Section~\ref{Sec:Stream}, the \textit{online generalization} and \textit{refinement} of PR-GPT shares a motivation similar to some streaming GP approaches. We demonstrate the potential of PR-GPT to support streaming GP by comparing its quality and efficiency with that of running the corresponding refinement method from scratch in each step.

As stated in Section~\ref{Sec:Prob}, the snowball model was adopted to simulate streaming GP, where we set the number of steps to $10$. The average evaluation results of streaming GP over five independent runs on the datasets with $N = 100$K are visualized in Fig.~\ref{Fig:Stream-GP}.
In addition, Table~\ref{Tab:Stream-N} reports the corresponding variation regarding (\romannumeral1) the number of nodes $N$ and (\romannumeral2) the average number of nodes $\tilde N$ in the initialization (i.e., the weighted super-graph) given by PR-GPT.

With the increase of step, the time of running a refinement method from scratch grows linearly. For PR-GPT, the increase of inference time is slightly sub-linear, which is consistent with the evaluation of streaming GP in some previous solutions \cite{uppal2018fast} submitted to Graph Challenge. PR-GPT can achieve quality close to the corresponding baselines in most steps and even obtain better quality in some cases.
As shown in Table~\ref{Tab:Stream-N}, PR-GPT significantly reduces the number of nodes to be processed. In particular, it can ultimately reduce the scale of a graph by about half as the step increases.
In summary, the pre-training \& refinement paradigm of PR-GPT has the potential to support streaming GP.

\section{Conclusion}\label{Sec:Conc}
In this paper, we considered $K$-agnostic GP and proposed PR-GPT. It follows a pre-training \& refinement paradigm, including the (\romannumeral1) \textit{offline pre-training} on historical small graphs as well as the (\romannumeral2) \textit{online generalization} to and (\romannumeral3) \textit{refinement} on new large graphs. We evaluated PR-GPT on the IEEE HPEC Graph Challenge benchmark, comparing its inference quality and efficiency over seven baselines.
Experiments demonstrated that PR-GPT, combined with different refinement methods (e.g., \textit{InfoMap} and \textit{Locale}), can achieve faster GP without significant quality degradation. Surprisingly, it can even achieve improvement for both quality and efficiency in some cases.
Based on a mechanism of providing a good initialization with a smaller scale (i.e., the number of nodes to be processed), PR-GPT also has the potential to support streaming GP and can obtain efficiency improvement consistent with the results of some previous work.

In this study, we only considered the snowball model of streaming GP. We plan to extend our method to another emerging edge model of Graph Challenge in our future work. Moreover, extending PR-GPT to (\romannumeral1) dynamic graphs \cite{lei2018adaptive,lei2019gcn,qin2023high,qin2023temporal} that has a motivation similar to streaming GP and (\romannumeral2) attributed graphs with the consideration of inherent correlations between graph topology and attributes \cite{qin2018adaptive,chunaev2020community,qin2021dual,zhao2022trade} are also our next research focuses.


\bibliographystyle{IEEEtran}

\end{document}